\documentclass[11pt]{article}

\usepackage[final]{acl}

\usepackage{times}
\usepackage{latexsym}

\usepackage[T1]{fontenc}

\usepackage[utf8]{inputenc}

\usepackage{microtype}

\usepackage{inconsolata}

\usepackage{graphicx}
\usepackage{amssymb} 
 \usepackage{amsmath}
  \usepackage{booktabs}
\usepackage{dsfont}
 
\usepackage{makecell}
\title{AVERT: Audio-Verified Adjudication for Spoken Dialogue State Tracking}

\author{Chunggi Lee \\
  Harvard University \\
  Cambridge, Massachusetts, USA\\
  \texttt{chunggi\_lee@g.harvard.edu} \\\And
  Hanspeter Pfister \\
  Harvard University \\
  Cambridge, Massachusetts, USA\\
  \texttt{pfister@g.harvard.edu} \\}

\begin{document}
\maketitle
\begin{abstract}
Spoken dialogue state tracking recovers slot-value pairs from speech, where ASR errors concentrate in entity values and persist across turns, making it both a generation and an editing problem. A strong per-turn text editor corrects much of this but, operating on the transcript alone, leaves three recoverable errors: a value predicted inconsistently across turns, an omitted slot, and a value the audio does not support. We present AVERT, which scores each candidate value by combining cross-turn agreement with a trained audio-conditioned verifier and resolves the three error types with three operators, vote, add, and swap, each restricted to the slots where its error is common. On SpokenWOZ, a base speech-LLM reaches 33.04 JGA, a text editor 38.34, and AVERT 40.13, without retraining either. This is in the range of a 1B end-to-end system that consumes the full spoken history (39.32), though AVERT uses two 1B decoders rather than one. The audio verifier contributes a statistically significant gain, and restricting each operator to a selected slot subset matters: removing it lets unrestricted voting overwrite correct categorical values and fall below the editor.
\end{abstract}

\section{Introduction}
Dialogue state tracking (DST) predicts a dialogue state, a set of slot-value pairs, from the dialogue history \citep{multiwoz,wu2019transferable,heck2020trippy}. Spoken DST must recover the same state from speech \citep{si2023spokenwoz}, which is harder because slot values must be inferred from noisy acoustic evidence. Most errors fall on entity values such as names, locations, and numbers, which ASR frequently mistranscribes or drops. In a deployed voice assistant, these turn directly into wrong actions. The state also accumulates across turns, so an early error persists and degrades later predictions. Spoken DST is therefore two problems at once: recovering the state from speech, and correcting the errors that build up.

Our pipeline has two existing components: a base spoken-DST model ($R$) that maps the current-turn audio and the text history to an ASR transcript and a first-pass state, and a per-turn text editor ($E$) \citep{tian2021agdst, li2023rspdst, he2024correctionlm} that revises that state from the cumulative transcript alone, without the audio. A strong $E$ already corrects much of the accumulated error. However, working from the transcript alone, $E$ leaves three recoverable errors: a value predicted inconsistently across turns, an omitted slot, and a value the audio does not support, such as a mistranscribed name. Two signals can address these. The first is cross-turn agreement, aggregated over the editor's and the base model's past predictions, which overrides a one-off inconsistent value \citep{wang2023selfconsistency}. But agreement has two blind spots. It cannot recover an omitted slot, since there is nothing to aggregate. And it cannot fix a value that the predictions agree on but the audio rejects. For those cases we need a signal that consults the audio directly, which we supply with a trained audio-conditioned verifier.

Existing systems do not couple cross-turn agreement with a value-level audio check. End-to-end speech-LLM systems take the audio as input \citep{elghazal2025lrec,vendrame2025arxiv}, but they never use it to verify individual candidate values. Correction and cascade systems work from the transcript alone \citep{he2024correctionlm,tian2021agdst,deragec} and have no access to the audio. In neither case does the audio adjudicate among competing candidates. Rather than retrain the base model to change how it consumes audio, we keep it fixed and train a separate verifier that operates over the editor's already-formed candidate values.

We propose AVERT (Audio-VERified adjudicaTion). AVERT trains an auxiliary verifier $\sigma$ that takes the turn audio, a slot, and a candidate value, and estimates whether the audio supports that value. For each candidate, $\sigma$ weights a cross-turn agreement count into a single score. A lexical attestation gate $\alpha$ then decides when a value may be inserted or substituted. It is a simple substring check against the cumulative ASR transcript. Three operators act on these scores. Vote selects the highest-scoring value. Add recovers an omitted slot when an attested, confident value exists. Swap replaces an unattested value with an attested alternative. We restrict each operator to the slots where its error is common, so every correction is tied to a specific slot and audio condition.

On SpokenWOZ \citep{si2023spokenwoz}, the base spoken-DST model $R$ reaches 33.04 JGA, the text-only editor $E$ 38.34, and AVERT 40.13, a $1.79$-point gain with no retraining of $R$ or $E$. This is in the range of the strongest comparable 1B configuration of \citet{elghazal2025lrec} (39.32), which feeds the entire spoken conversation to the LLM, though AVERT uses two 1B decoders rather than one. The difference is where audio enters: AVERT keeps the history as text and uses audio only in a lightweight per-turn verifier over candidate values, so its audio context does not grow with dialogue length and its gain over the editor rises from $+0.6\%$ on turns 1--5 to $+19.4\%$ on turns 31+ (\S\ref{sec:analysis}).

We make two contributions. First, we introduce a value-level adjudication mechanism combining cross-turn agreement, a trained audio verifier, and lexical attestation. Without retraining $R$ or $E$, it lifts a strong 1B editor by $1.79$ JGA into the range of 1B systems that read the full spoken history, at roughly twice the decoder parameters. Two controls show that simply merging $R$'s predictions into $E$ does not produce this gain (\S\ref{sec:analysis}). Second, restricting each operator to a selected slot subset matters: voting over all slots overwrites correct categorical and boolean values (114 to 675 changes), dropping JGA by up to 3.6 points, below the editor.

\section{Related Work}
\subsection{Spoken Dialogue State Tracking}
Spoken DST predates the current speech-LLM line. The early Dialog State Tracking Challenges \citep{williams2013dialog, henderson2014second, henderson2014third} tracked state over ASR $N$-best lists with confidence scores, and DSTC11 revisited the setting with a speech-aware task-oriented dialogue track \citep{soltau2023dstc}. These systems consumed ASR output rather than audio, so the acoustic signal was available only through recognizer confidence, never as evidence for a specific slot value.
\citet{si2023spokenwoz} released SpokenWOZ and showed that text DST methods transfer poorly to spoken data, a gap that has since been studied through targeted analysis \citep{druart2023}. A recent line of work addresses it with end-to-end speech-LLM models that take the audio as input. \citet{sedlavcek2025interspeech} align WavLM-large \citep{chen2022wavlm} with a small connector and an OLMo-1B or Gemma-2-9B decoder. \citet{elghazal2025lrec} feed the full or compressed spoken history to the LLM. \citet{vendrame2025arxiv} train jointly on spoken and out-of-domain textual DST data, and \citet{gulzar2025} transform written DST data into a spoken style, both for better generalization. On the data side, \citet{lee2023exploring} synthesize audio from written DST corpora with TTS and study how far such synthetic speech can substitute for real recordings when training audio-based DST models. These approaches address the scarcity of spoken DST data by enlarging or converting the training set, whereas AVERT leaves the training data and the base model untouched and adjudicates the entity values a text editor has already produced. \citet{ka2g} instead ground slot-value prediction in the audio, but do so inside a trained pointer-generator that produces the state, so the audio is internal to the model and cannot be inspected per value. We adopt the WavLM-plus-OLMo backbone of \citet{sedlavcek2025interspeech} as our base spoken-DST model and keep it fixed. Unlike all of the above, AVERT does not use audio to generate the dialogue state at all: it trains a separate verifier that scores already-formed entity values against the turn audio.

\subsection{Self-Correction and Candidate Re-ranking}
A separate line of work revises predictions after a first pass. In text DST, amendable generation \citep{tian2021agdst}, revisable state prediction \citep{li2023rspdst}, structural state correction \citep{su2023scalabledsc}, and self-correction with small models \citep{he2024correctionlm} all edit an initial state, but operate on the text alone. Self-consistency \citep{wang2023selfconsistency} selects an answer by majority vote over multiple samples. We aggregate analogously across past turns and prediction sources, but instead of relying on agreement frequency alone we weight each vote by a trained audio verifier. Closest to our mechanism, ClozeGER \citep{clozeger} reformulates ASR error correction as a cloze test in which a multimodal model chooses among N-best candidates conditioned on the source speech, and DeRAGEC \citep{deragec} denoises named-entity candidates before any state is formed. Both operate at the transcript level and produce text. In particular, ClozeGER chooses among ASR N-best hypotheses to emit a corrected transcript before any state is built, so its audio conditioning targets words, whereas AVERT conditions on the queried slot and re-ranks slot values already placed in the structured state, over the editor's outputs and without retraining the base, so each correction is tied to a specific slot and audio condition.
\section{Method}
\begin{figure}[t]
\centering
\includegraphics[width=\linewidth]{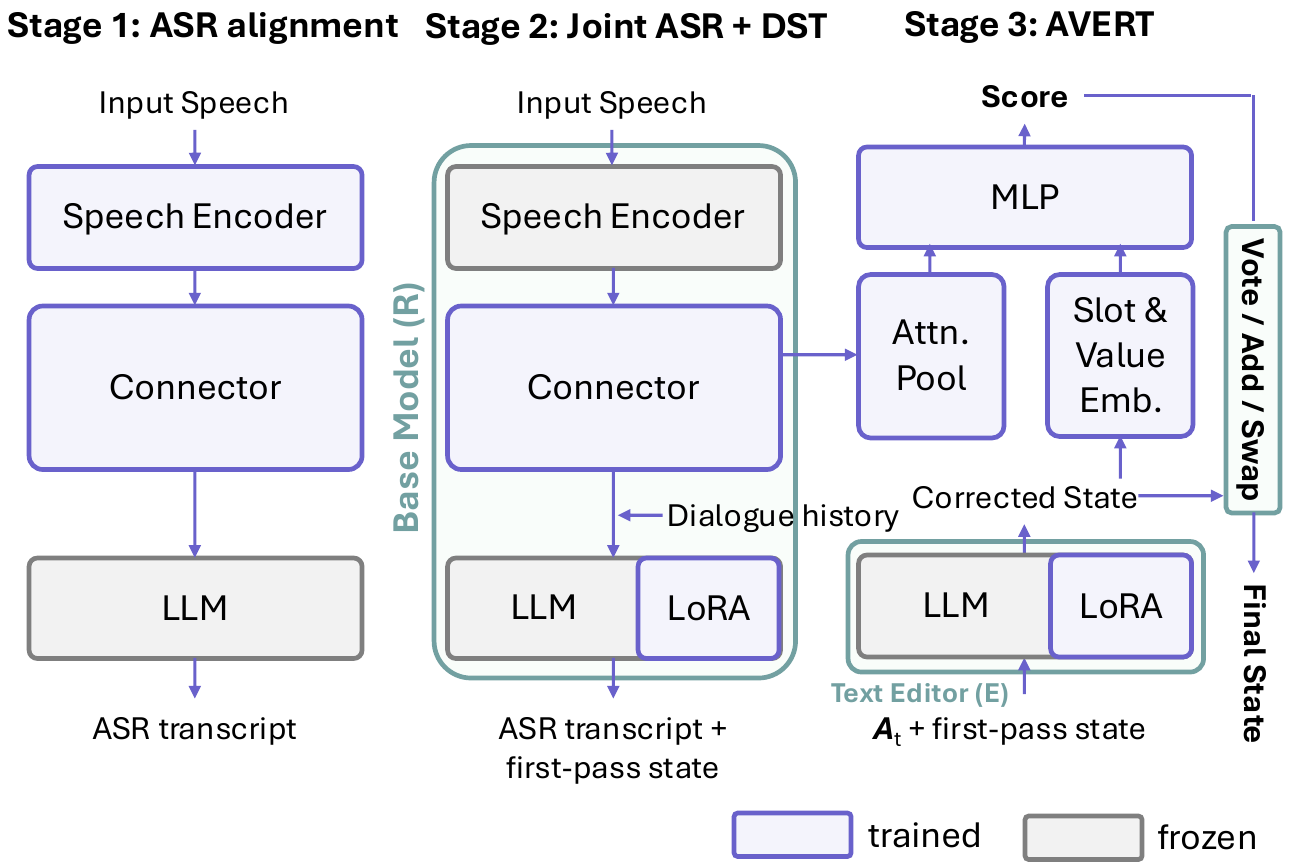}
\caption{Component view of the pipeline, with trained and frozen modules marked. The base spoken-DST model $R$ is trained in two stages: the speech encoder and connector are first aligned for ASR with the LLM frozen, then the connector and LoRA adapters are fine-tuned for joint ASR and DST with the encoder frozen. At inference $R$ takes the current-turn audio and the dialogue history as text, and emits an ASR transcript together with a first-pass state. The text editor $E$ is a separately LoRA-tuned decoder that maps the cumulative transcript $A_t$ and the first-pass state to a corrected state, and never receives the audio. The verifier $\sigma$ is the only module AVERT adds: it reuses $R$'s frozen encoder and connector, and trains an attention pool, the slot and value embeddings, and an MLP that scores a candidate value against the turn audio. The three operators use these scores to edit the corrected state into the final state.}
\label{fig:architecture}
\end{figure}

\begin{figure*}[t]
\centering
\includegraphics[width=\linewidth]{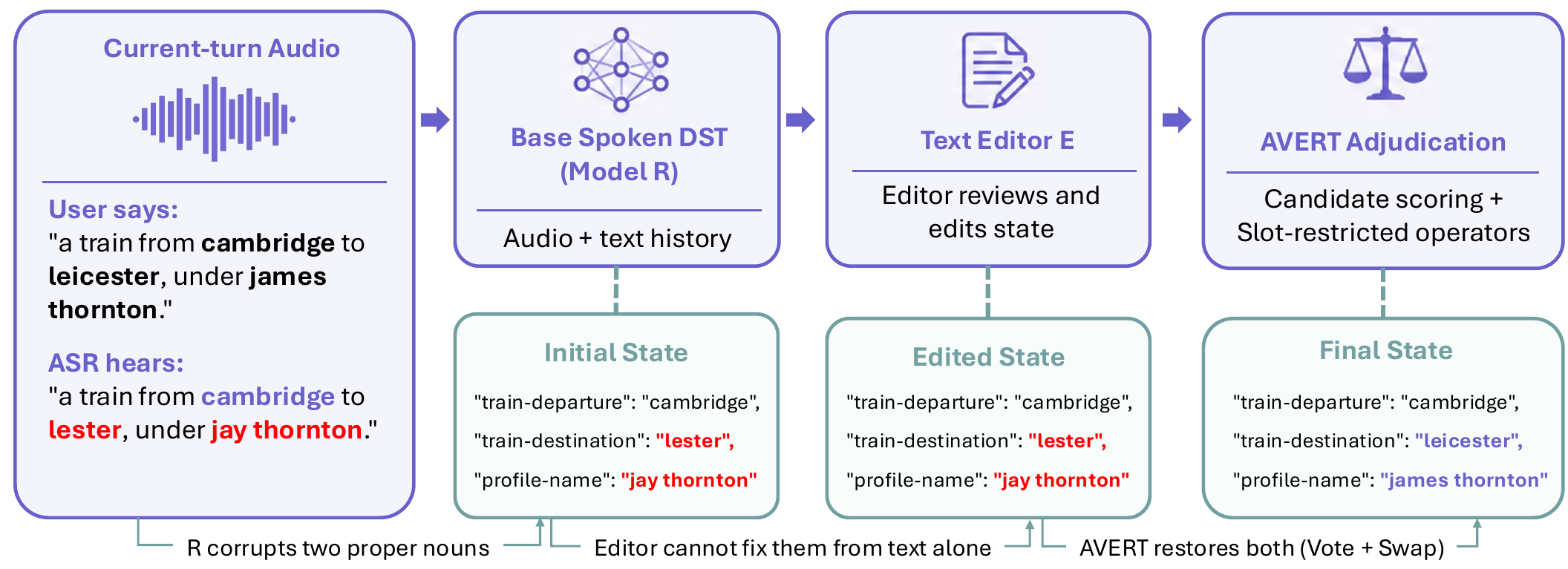}
\caption{Overview of AVERT. The base spoken-DST model $R$ emits a transcript and a first-pass state, the text editor $E$ revises it, and AVERT adjudicates candidate values across turns with three slot-restricted operators. Here ASR corrupts two proper nouns (\textit{leicester} to \textit{lester}, \textit{james} to \textit{jay}), which the editor cannot fix from the transcript alone. AVERT restores both: VOTE recovers the cross-turn consensus \textit{leicester}, and SWAP replaces the unattested \textit{jay thornton} with an attested alternative, \textit{james thornton}.}
\label{fig:overview}
\end{figure*}

Figure~\ref{fig:architecture} shows which modules are trained and which are frozen, and Figure~\ref{fig:overview} walks through the full pipeline on a running example: the base model $R$ produces a first-pass state, the editor $E$ revises it, and AVERT adjudicates the candidate values.

\subsection{Task and Notation}
A dialogue $d$ is a sequence of turns. In spoken DST the input at turn $t$ is the audio rather than the transcript, and the goal is to predict the cumulative state $B_t$, a set of slot-value pairs $(s,v)$ over a schema of slots $S$ \citep{si2023spokenwoz}. We evaluate with joint goal accuracy (JGA), the fraction of turns whose predicted state matches the gold state exactly, with no missing or extra pair. Following \citet{sedlavcek2025interspeech}, named-entity slots are matched by normalized Levenshtein similarity at a threshold of $0.90$ and all other slots by string equality. Following prior work \citep{sedlavcek2025interspeech,elghazal2025lrec} we use causal per-turn evaluation: each turn is predicted from the model's own previous outputs, so errors propagate. We write $\hat v(d,t,s)$ for the value predicted for slot $s$ at turn $t$ of dialogue $d$.

\subsection{Base Model and Editor}
\textbf{Base spoken-DST model.} Figure~\ref{fig:architecture} shows the components and which parts of each are trained. We build on the speech-LLM design for spoken DST \citep{sedlavcek2025interspeech, elghazal2025lrec} and keep this backbone rather than a stronger encoder, so the reported differences reflect AVERT rather than a change of base architecture. The base model $R$ maps the user's speech to a structured output at every turn and has three parts: a WavLM-large encoder \citep{chen2022wavlm} that encodes the turn audio into frame-level representations, a lightweight connector projecting these into the LLM embedding space, and an OLMo-2-1B \citep{olmo2} adapted with LoRA \citep{hu2022lora}. $R$ is trained in two stages. First the encoder and connector are trained for ASR with the LLM frozen, then the connector and LoRA adapters are fine-tuned on SpokenWOZ for joint ASR and DST with the encoder frozen. Conditioned on the current-turn audio and the dialogue history as text, $R$ emits at each turn a JSON object with the ASR transcript and a first-pass state. We write $A_t = a_1 \oplus \cdots \oplus a_t$ for the cumulative transcript, where $a_j$ is the transcript $R$ produces at turn $j$.

\textbf{Text editor.} On top of $R$, a per-turn text editor $E$ revises the first-pass state. $E$ is a separate OLMo-2-1B \citep{olmo2} with LoRA \citep{hu2022lora}, fine-tuned once on SpokenWOZ to map $A_t$ and the first-pass state to a corrected state. We use $R$ and $E$ as the two prediction sources for AVERT. The editor baseline applies $E$ alone, $\tilde{v}_{\mathrm{base}}(d,t,s)=\hat{v}_E(d,t,s)$, which AVERT refines. Because it conditions on the full cumulative transcript $A_t$, $E$ resolves most text-evident errors, which is why it already improves substantially over $R$. The errors it cannot resolve from text alone are precisely the three AVERT targets.

\subsection{Audio as a Verifier, not a Generator}
A central design choice in AVERT is to use the audio to verify candidate values rather than to regenerate the state. Once $R$ has produced the transcript $A_t$ and a first-pass state, deciding which slots to keep and how to reconcile their values across turns is largely a structural operation over text, which a text editor handles well. The distinctive contribution of the audio lies elsewhere, at the level of individual entity values, where it can confirm or reject a specific candidate that the transcript may have corrupted. Verifying whether a proposed value is supported by the audio is a narrower and more reliable use of the signal than generating the entire state from it.  We therefore keep the editor text-only and train a separate audio verifier that scores its outputs at the value level, adjudicating among already-formed candidate values rather than producing them from scratch.

\begin{figure*}[t]
    \centering
    \includegraphics[width=\textwidth]{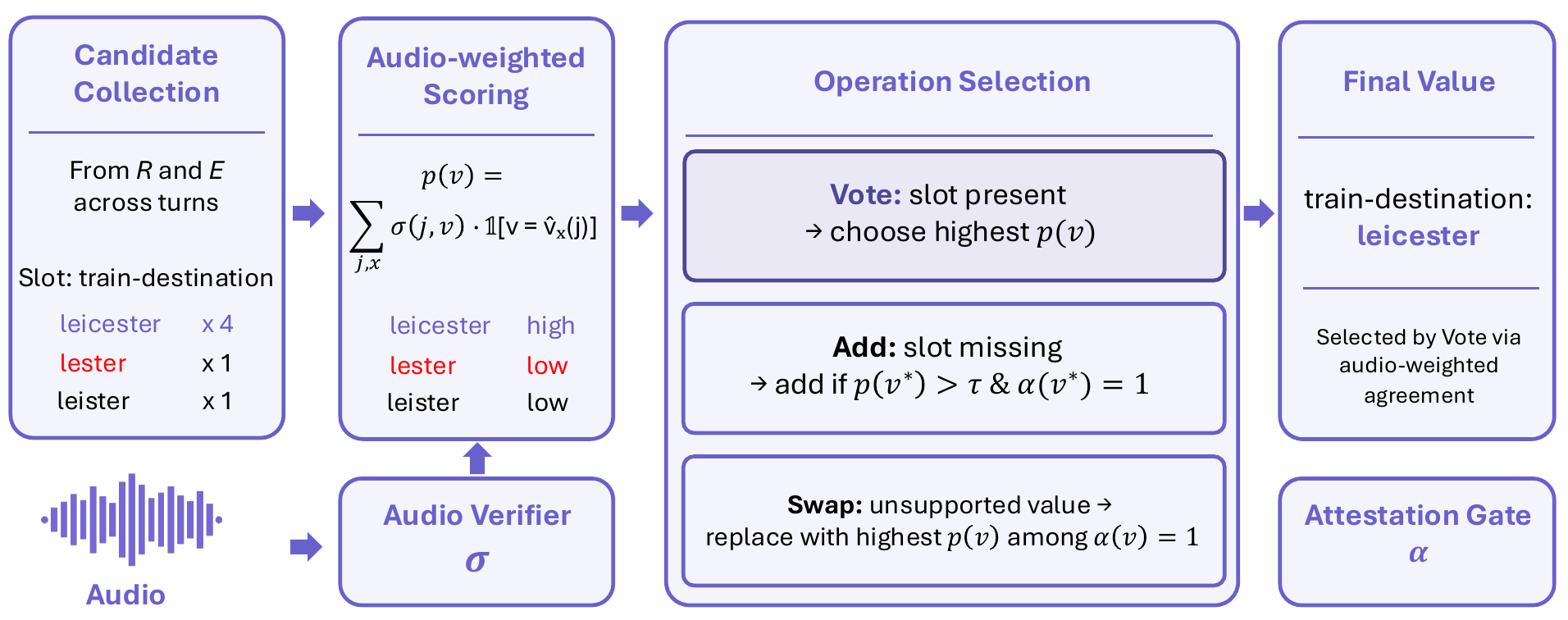}
\caption{AVERT adjudication for the running example. Candidate values for the target slot are collected from $R$ and $E$ across turns and scored by audio-weighted agreement using the trained verifier $\sigma$. Because the slot is already present, AVERT applies \textsc{Vote} and selects the highest-scoring candidate, \textit{leicester}. The attestation signal $\alpha(v)$ constrains only \textsc{Add} and \textsc{Swap}. Counts such as x4 are the number of (source, turn) predictions supporting that value. A repeated value is stored once in $V$ but each occurrence is counted.}
\label{fig:avert-adjudication}
\end{figure*}

\subsection{Candidate Collection and Scoring}
\label{sec:scoring}
Figure~\ref{fig:avert-adjudication} walks through this scoring and the operator selection on the running example.
For slot $s$ at turn $t$ of dialogue $d$, AVERT scores each candidate value using evidence aggregated across the two sources and all past turns. We gather the candidates $\hat{v}_x(j)$ for $x \in \{E, R\}$ and $j \leq t$, writing $\hat{v}_x(j) = \varnothing$ when $x$ does not predict $s$ at $j$, and let $V$ be the set of distinct non-empty values among them. A value that several sources or turns predict appears in $V$ once, but every one of its occurrences is counted in the score below. Each candidate $v \in V$ receives an agreement score that counts the (source, turn) pairs predicting it, and a trained audio-conditioned confidence $\sigma(j,v) \in [0,1]$ that estimates whether the audio of turn $j$ supports $v$ (\S\ref{sec:classifier}):
\begin{equation}
A(v) = \sum_{j \le t} \sum_{x \in \{E,R\}} \sigma(j,v)\, \mathbf{1}[v = \hat v_x(j)].
\end{equation}
The candidate score is this audio-weighted agreement directly,
\begin{equation}
p(v) = A(v).
\end{equation}
We additionally define a lexical attestation indicator $\alpha(v)=\mathbf{1}[\mathrm{first\text{-}token}(v)\in A_t]$, a substring check against the cumulative ASR transcript. We use first-token attestation because ASR often corrupts multi-token entity values only partially, so a full-string match would reject many recoverable values. Unlike $\sigma$, $\alpha$ does not consult the audio: it is a transcript-level signal. Since $\alpha$ acts only as a gate for \textsc{Add} and \textsc{Swap} (\S\ref{sec:operators}) and not as a ranking score, an attested candidate is selected only when it has sufficient support under $p(v)$. Ties are rare for the same reason: $p(v)$ weights each occurrence by $\sigma$ rather than counting occurrences uniformly, so candidates with equal raw frequency can still be separated.

\subsection{Audio-Conditioned Verifier}
\label{sec:classifier}
To verify a value against the audio, AVERT trains a classifier $\sigma(j,v) \in [0,1]$ that estimates whether the audio of turn $j$ supports value $v$ for slot $s$. It is conditioned on three inputs, the audio, the value, and the slot. The slot is necessary because the same utterance can support different values depending on the queried slot. As the feature extractor we reuse $R$'s frozen speech encoder and connector. The turn-$j$ audio yields frame embeddings $\{h_i\}$, which a single learned-query attention pool compresses into an audio vector $u_j$,
\begin{equation}
u_j = \sum_i a_i h_i, \qquad a_i = \frac{\exp(q^\top h_i)}{\sum_{i'} \exp(q^\top h_{i'})},
\end{equation}
where $q$ is a learned query. The value $v$ is tokenized with $R$'s tokenizer and mean-pooled into $e_v$, and the slot maps through an embedding table to $e_s$. A three-layer MLP with GELU activations \citep{hendrycks2016gelu} and dropout maps the concatenation to the support score,
\begin{equation}
\sigma(j,v) = \mathrm{sigmoid}\,\mathrm{MLP}([\,u_j;\, e_s;\, e_v\,]).
\end{equation}
We train $\sigma$ on SpokenWOZ by sampling $K{=}8$ candidate values per (turn, slot) from the editor $E$, labeling each positive if it matches the gold value and negative otherwise. As the samples are negative-skewed, we use focal binary cross-entropy \citep{lin2017focal}. Only the attention pool, the embedding tables, and the MLP are trained, so $\sigma$ is the single trained component AVERT introduces beyond the existing $R$ and $E$. At inference, $\sigma$ scores candidates from both $E$ and $R$, which share the SpokenWOZ value space, so it transfers across sources without source-specific training.

\subsection{Operators}
\label{sec:operators}
Let $v_E$ be the editor's value for slot $s$ at the current turn, $v^\star = \arg\max_{v \in V} p(v)$ the highest-scoring candidate, and $\hat v$ the current value, initialized to $v_E$. AVERT applies three operators in sequence, each enabled only on a slot subset selected on the development set. For each operator we include a slot only when enabling it corrects more errors than it introduces on the development set. The subsets are fixed before any test evaluation and listed in Appendix~\ref{app:slots}.

\textbf{Vote.} If the slot is present ($v_E \ne \varnothing$), replace it with the consensus value, $\hat v \leftarrow v^\star$, resolving values predicted inconsistently across turns.

\textbf{Add.} If the slot is omitted ($v_E = \varnothing$), insert $v^\star$ when it is attested and confident, $\alpha(v^\star) = 1$ and $p(v^\star) > \tau$, setting $\hat v \leftarrow v^\star$. 

\textbf{Swap.} If the current value is present but unattested ($\hat v \ne \varnothing$ and $\alpha(\hat v) = 0$), replace it with the highest-scoring attested alternative, $\hat v \leftarrow \arg\max_{v \in V:\, \alpha(v)=1} p(v)$, when one exists.

AVERT adds no trained parameters beyond $\sigma$. The add threshold $\tau$ is tuned on the development set and its value is given in \S\ref{sec:impl}.

\subsection{Implementation Details}
\label{sec:impl}
The verifier $\sigma$ is a three-layer MLP with GELU activations \citep{hendrycks2016gelu} over the pooled turn-$j$ frame representations, the slot embedding, and the mean-pooled candidate-value embedding, using dropout $0.1$ and hidden size $512$, trained with focal loss \citep{lin2017focal} ($\gamma = 2.0$). The add threshold is $\tau = 0.3$, tuned on the development set. Stage 1 training uses four NVIDIA A100 GPUs, while Stage 2 fine-tuning and verifier training use one.
\section{Experiments}
\subsection{Setup}
\label{sec:setup}
We evaluate on the SpokenWOZ test set \citep{si2023spokenwoz} under strictly causal per-turn evaluation, reporting JGA with fuzzy matching on named-entity slots \citep{sedlavcek2025interspeech}. The base model $R$ is trained in two stages: Stage 1 aligns a WavLM-large encoder \citep{chen2022wavlm} and a connector into a frozen OLMo-2-1B \citep{olmo2} for ASR on roughly $4{,}600$ hours (SpokenWOZ, Loquacious \citep{loquacious2024}, and Fisher \citep{cieri2004fisher}), and Stage 2 adds a rank-16 LoRA adapter \citep{hu2022lora} and fine-tunes on SpokenWOZ for joint DST and ASR. The editor $E$ is a separate rank-16 LoRA-tuned OLMo-2-1B trained on (R prediction, gold state) pairs, and the verifier $\sigma$ is trained on $K{=}8$ candidate values per (turn, slot) sampled from $E$ with focal binary cross-entropy. Remaining configuration is in \S\ref{sec:impl}.

\paragraph{Inference cost.} On a single A100 with batch size 1, $R$ takes 2{,}558\,ms per turn at 11.76\,GB peak memory, $E$ takes 1{,}534\,ms at 5.52\,GB, and the verifier takes 56\,ms at 6.23\,GB, which includes the frozen encoder and connector it reuses from $R$. The full pipeline runs in 4{,}148\,ms per turn, and peak memory stays near 11.8\,GB because the stages run sequentially rather than concurrently. The verifier is about 2\% of the base model's latency, since it runs a small MLP over representations the encoder has already produced.

\subsection{Main Results}
\label{sec:main_results}
Table~\ref{tab:main} compares AVERT against published causal baselines among 1B-class systems. Our pipeline improves from a 33.04 first pass by $R$ to 38.34 after the editor $E$, and AVERT raises this to 40.13 ($+1.79$ over the editor), without retraining $R$ or $E$. \citet{elghazal2025lrec} reach 39.32 by feeding the entire spoken conversation to the LLM, at an audio-context cost that grows with dialogue length. AVERT reaches 40.13 with two 1B decoders against their one, so the two are not matched in capacity, but it keeps the dialogue history as text and leaves the audio context constant in turn count. Even without the verifier it reaches 39.74 (Table~\ref{tab:ablation}), and the verifier adds a further significant 0.40 points.

Because AVERT's add and swap operators edit values directly, we also report exact matching (Table~\ref{tab:matching}). The gain over the editor holds under both metrics, reaching $+1.54$ under exact and $+1.79$ under fuzzy, so AVERT recovers exactly correct values rather than fuzzy-close ones. Fuzzy is the metric comparable to the baselines in Table~\ref{tab:main}.

The lower block of Table~\ref{tab:main} lists 9B-backbone systems for context. These score higher than AVERT but use roughly an order of magnitude more decoder parameters, so we scope our claim to the 1B regime and do not treat them as comparison points.

\begin{table}[t]
\centering
\small
\begin{tabular}{lc}
\toprule
Model & JGA \\
\midrule
\multicolumn{2}{l}{\emph{1B-class:}}\\
SPACE+WavLM \citep{si2023spokenwoz} & 25.65 \\
Whisper+T5 \citep{druart2023} & 24.10 \\
UBAR+GenWOZ \citep{gulzar2025} & 25.90 \\
\citet{sedlavcek2025interspeech} (1B) & 34.66 \\
\citet{elghazal2025lrec}, comp.\ (1B) & 36.49 \\
\citet{elghazal2025lrec}, full (1B) & 39.32 \\
\midrule
\multicolumn{2}{l}{\emph{Ours:}}\\
Base $R$ (first pass) & 33.04 \\
Editor $E$ & 38.34 \\
$E$ + AVERT & \textbf{40.13} \\
\midrule
\multicolumn{2}{l}{\emph{Larger backbones (9B, not matched):}}\\
\citet{sedlavcek2025interspeech} (9B) & 42.17 \\
\citet{elghazal2025lrec}, comp.\ (9B) & 43.16 \\
\citet{elghazal2025lrec}, full (9B) & 45.52 \\
\bottomrule
\end{tabular}
\caption{JGA (\%) on SpokenWOZ test, strictly causal. Among 1B-class systems $E$+AVERT is highest without retraining the base. AVERT is a pipeline of a 1B base $R$ and a 1B editor $E$ with a small verifier (roughly twice the decoder parameters of a single-model 1B system), compared here to single-model 1B-class systems. }
\label{tab:main}
\end{table}

\begin{table}[t]
\centering
\small
\begin{tabular}{lcc}
\toprule
Variant & JGA (w/o fuzzy) & JGA (w/ fuzzy) \\
\midrule
Base $R$ (first pass) & 31.73 & 33.04 \\
Editor $E$ & 36.12 & 38.34 \\
$E$ + AVERT & \textbf{37.66} & \textbf{40.13} \\
$\Delta$ (AVERT $-$ $E$) & $+1.54$ & $+1.79$ \\
\bottomrule
\end{tabular}
\caption{The gain over the editor holds under both exact (w/o fuzzy) and fuzzy (w/ fuzzy, Levenshtein 0.90) matching, so it is not driven by the matching tolerance. Fuzzy is the metric comparable to the baselines in Table~\ref{tab:main}.}
\label{tab:matching}
\end{table}

\subsection{Ablation}
Table~\ref{tab:ablation} ablates AVERT along two axes: removing each scoring signal from the full model, and enabling the three operators cumulatively on E.

\paragraph{Audio verifier.} Removing $\sigma$ from the score, leaving uniform-weighted agreement, drops AVERT from 40.13 to 39.74. A paired bootstrap ($B=1000$, dialogue resampling) puts this gain at $+0.40$ (95\% CI $[+0.08, +0.71]$, $p=0.016$). The same test inside the vote operator alone gives $+0.34$ ($p=0.024$), so the verifier's contribution is concentrated in vote. To check that $\sigma$ relies on the audio rather than on the slot and value alone, we zero the pooled audio representation while keeping the other two inputs unchanged: JGA falls to 39.80, only 0.06 above the no-$\sigma$ configuration, so the audio input accounts for most of the verifier's gain.

\paragraph{Attestation gate.} The attestation indicator $\alpha$ enters AVERT only as a gate on the add and swap operators, where it is load-bearing: removing the in-audio constraint drops JGA to 38.95, a 1.18-point loss ($p < 10^{-4}$).

\paragraph{Operators.} All scoring signals are active throughout this cascade, and operators are enabled cumulatively on top of the editor. Vote alone reaches 39.42, adding add brings it to 39.97, and swap reaches 40.13. The marginal contribution of swap over vote and add is $+0.16$ (95\% CI $[+0.02, +0.36]$, $p=0.014$), small but significant. Vote is the principal operator, reflecting that the dominant recoverable error in a strong per-turn editor is cross-turn inconsistency rather than omission.

\begin{table}[t]
\centering
\small
\begin{tabular}{lcc}
\toprule
Variant & JGA & $\Delta$ \\
\midrule
$E$ baseline (no AVERT) & 38.34 & --- \\
AVERT (full) & \textbf{40.13} & $+1.79$ \\
\midrule
\multicolumn{3}{l}{\emph{Signal ablations (vs.\ full AVERT):}}\\
without $\sigma$ (uniform agreement) & 39.74 & $-0.40$ \\
\ $\sigma$ with audio zeroed & 39.80 & $-0.33$ \\
without $\alpha$ gate (no in-audio) & 38.95 & $-1.18$ \\
\midrule
\multicolumn{3}{l}{\emph{Operators (cumulative, vs.\ editor):}}\\
\ + Vote & 39.42 & $+1.07$ \\
\ + Vote + Add & 39.97 & $+1.62$ \\
\ + Vote + Add + Swap & 40.13 & $+1.79$ \\
\bottomrule
\end{tabular}
\caption{AVERT ablation on SpokenWOZ test. Signal-ablation $\Delta$ is relative to full AVERT (40.13); cumulative operator $\Delta$ is relative to the editor $E$ (38.34) with all signals active. Operator 95\% CIs from a paired dialogue-level bootstrap are $[+0.71, +1.46]$, $[+1.17, +2.11]$, and $[+1.28, +2.32]$.}
\label{tab:ablation}
\end{table}

\subsection{Effect of Slot Restriction}
\label{sec:perslot}
Each operator is enabled only on a slot subset selected on the development set. To test whether this restriction matters, we remove it. B1 applies $\sigma$-only add and swap to all slots with a single global threshold and no editor-anchored vote, and B2 further adds an unrestricted cross-turn vote over all slots. At dev-selected thresholds, B1 reaches 38.14 ($-1.99$, $p<10^{-4}$) and B2 36.53 ($-3.60$, $p<10^{-4}$), neither robustly above the editor. The swap counts are not comparable across rows. SWAP fires only on present but unattested values, so without an editor-anchored vote the current value stays the editor's own, which is usually attested. ADD depends only on the threshold. B2's failure is visible in the edit counters (Table~\ref{tab:perslot}). Unrestricted voting overwrites correct categorical and boolean values, raising such changes from 114 to 675 and swaps from 2{,}501 to 7{,}359, since a boolean slot like parking has no useful cross-turn entity signal. The operators thus work best when restricted to their selected slots rather than applied globally. Generating the state from an empty start with $\sigma$ alone reaches only 35.61, confirming that anchoring on the editor is essential.

\begin{table}[t]
\centering
\small
\begin{tabular}{lcccc}
\toprule
Variant & JGA & added & swapped & cat/bool chg.\\
\midrule
AVERT & 40.13 & 1{,}851 & 2{,}501 & 114 \\
B1 & 38.14 & 3{,}011 & 114 & 34 \\
B2 & 36.53 & 3{,}016 & 7{,}359 & 675 \\
\bottomrule
\end{tabular}
\caption{Removing slot restriction (B1, B2) underperforms AVERT at every threshold. Thresholds for B1 and B2 are selected on the development set. Unrestricted cross-turn voting (B2) overwrites correct categorical and boolean values (114 $\rightarrow$ 675 changes), the principal cause of its drop. Edit counts are not comparable across rows, since each operator's firing condition depends on which operators precede it.}
\label{tab:perslot}
\end{table}

\begin{table}[t]
\centering
\small
\begin{tabular}{lccc}
\toprule
Defect type & $E$ & AVERT & $\Delta$ \\
\midrule
Omission            & 9{,}181 & 7{,}693 & $-16.2\%$ \\
Inconsistency       & 7{,}109 & 5{,}916 & $-16.8\%$ \\
Hallucination       & 5{,}834 & 6{,}197 & $+6.2\%$ \\
Other wrong value   & 2{,}415 & 2{,}984 & $+23.6\%$ \\
\midrule
Total               & 24{,}539 & 22{,}790 & $-7.1\%$ \\
\quad of which attested & 5{,}148 & 3{,}738 & $-27.4\%$ \\
\bottomrule
\end{tabular}
\caption{Slot-level error counts before and after AVERT (test, causal). The four types partition all wrong slots. Attested errors are those whose reference value appears in the cumulative self-ASR transcript, under a stricter substring criterion than the $\alpha$ gate of \S\ref{sec:operators}; AVERT repairs 1{,}435 of the 5{,}148 attested errors (27.9\%) and introduces 25.}
\label{tab:defects}
\end{table}

\section{Analysis}
\label{sec:analysis}

\paragraph{Which defects does AVERT fix?} We classify slot-level errors by automatic predicates over the editor output, gold state, and candidate history (Table~\ref{tab:defects}). Omission and inconsistency both fall by about 16\%, at the cost of a 6.2\% rise in hallucinations and a 23.6\% rise in other wrong values from \textsc{Add}'s insertions; the net effect is the 1.79-point gain of Table~\ref{tab:ablation}. Of the editor's 24{,}539 wrong slots, only 21.0\% are attested in the cumulative transcript, and AVERT repairs 27.9\% of that ceiling while introducing 25 new attested errors. The remaining 79.0\% would need stronger phonetic or generative grounding. Both fractions depend on the ASR error profile of SpokenWOZ and on what a 1B editor leaves behind.

\paragraph{Value errors or slot-presence errors?} AVERT's operators act on candidate values, but \textsc{Add} changes which slots are present at all, so we separate the two. Of the 2,506 slot-level errors AVERT corrects, 57.5\% are wrong values and 42.5\% are missing or spurious slots. Slot-presence recall rises from 89.66 to 91.22 and precision falls from 94.51 to 93.82, for an F1 of 92.50 against the editor's 92.02. The precision loss is the same effect as the hallucination increase in Table~\ref{tab:defects}: \textsc{Add} inserts slots the editor had left out, and not all of them are correct.

\begin{table*}[t]
\centering
\footnotesize
\setlength{\tabcolsep}{4pt}
\hyphenpenalty=10000 \exhyphenpenalty=10000
\begin{tabular}{@{}l
  >{\raggedright\arraybackslash}p{2.5cm}
  >{\raggedright\arraybackslash}p{3.9cm}
  >{\raggedright\arraybackslash}p{1.8cm}
  >{\raggedright\arraybackslash}p{2.7cm}
  >{\raggedright\arraybackslash}p{2.0cm}@{}}
\toprule
Operator & Slot (gold) & ASR transcript (excerpt) & $V$ & Signals fired & AVERT \\
\midrule
\textsc{Vote} & train-destination (cambridge) & Earlier turns repeatedly mention \emph{cambridge}. Current turn is a thank-you with no destination cue. & cambridge, melbourne & cambridge agreed in 32 of 34 past turns & cambridge \\
\addlinespace[2pt]
\textsc{Add} & restaurant-name (anatolia) & ``...moderate price range in the center. the anatolia, please.'' & (omitted) & $\sigma{=}0.72$ on anatolia; multi-source agree; $\alpha{=}1$ & anatolia \\
\addlinespace[2pt]
\textsc{Swap} & profile-name (james thornton) & ``james thornton.'' / ``okay, james thornton...'' & jay thornton (not in $A_t$) & $\alpha(\text{jay thornton}){=}0$; james thornton attested earlier & james thornton \\
\bottomrule
\end{tabular}
\caption{Qualitative traces from the SpokenWOZ test set, one per operator.}
\label{tab:traces}
\vspace{-1em}
\end{table*}

\paragraph{Does the second decoder explain the gain?} AVERT draws candidates from both $R$ and $E$, so we first test whether merging the two is enough. A naive union that adds every $R$-only slot to $E$ reaches 37.34, and a cross-turn consistency filter that adds an $R$-only slot only when $R$ predicts the same value earlier reaches 37.93. Both fall below $E$'s 38.34. The union restores false-positive slots that $E$ had removed, and the filter suppresses one-off noise but cannot correct a value $R$ predicts consistently wrong. All three configurations use the same two decoders, so the difference comes from how the predictions are adjudicated rather than from having them.


\textbf{Gains by slot type and within vote.} AVERT improves all three slot categories, with the largest absolute gain on numeric ($+2.51$), then categorical ($+1.61$) and proper-noun ($+1.54$). The proper-noun gain is smaller because that category has the lowest base accuracy and largest candidate space, where attestation is hardest. Within vote, uniform-weighted agreement gives $+0.73$ and weighting by $\sigma$ adds a further $+0.34$, so the verifier accounts for about a third of vote's gain and a fifth of the total improvement, neither hidden nor overstated.
The verifier also settles ties: among the $7{,}760$ \textsc{Vote} decisions with at least two distinct candidates, $331$ ($4.3\%$) tie on raw frequency, and the audio-weighted score separates every one of them.

\textbf{Gains grow with dialogue length.} AVERT's relative improvement rises monotonically with turn position, from $+0.6\%$ on turns 1--5 to $+19.4\%$ on turns 31+, as the cross-turn evidence accumulates. Long dialogues, where the editor is weakest (11.92\% accuracy at turn 31+), benefit most, and this is also where the full-spoken-history alternative is most expensive, so AVERT's advantage compounds.

\textbf{Qualitative cases.} Table~\ref{tab:traces} shows one trace per operator. Vote restores \emph{cambridge} against an isolated \emph{melbourne} drift backed by 32 of 34 past turns. Add recovers an omitted slot when the verifier proposes a multi-source-supported, attested value ($\sigma{=}0.72$ for \emph{anatolia}). Swap replaces the unattested misspelling \emph{jay thornton} with the attested \emph{james thornton} from earlier turns.
\section{Conclusion}
We presented AVERT, a value-level adjudication mechanism for spoken dialogue state tracking that combines cross-turn agreement with a trained audio-conditioned verifier and repairs three recoverable editor errors with three slot-restricted operators. On SpokenWOZ under strictly causal evaluation, it raises a 1B text editor from 38.34 to 40.13 JGA without retraining $R$ or $E$. Because audio enters only through a per-turn verifier, the language-model context does not grow with dialogue length, so AVERT is cheapest on the long dialogues where reading the full spoken history is most expensive. Slot restriction is also essential: unrestricted voting overwrites correct categorical values and falls below the editor.

\section*{Limitations}
\paragraph{Scale.} We test only the 1B backbone and do not run 9B or larger backbones. We expect the editor delta to transfer, since neither the editor nor the verifier depends on backbone size in any obvious way, but this remains a hypothesis.

\paragraph{Lexical attestation.} The attestation gate uses a literal first-token match against the cumulative transcript. A consistently mistranscribed value can therefore fail to be attested even when phonetically close to the audio. A soft phonetic attestation, for example via articulatory-feature similarity as in \citet{deragec}, is left to future work.


\paragraph{Attestation rate.} Only 21.0\% of the editor's errors have their reference value attested in the cumulative ASR transcript. The remaining 79.0\% are not reachable by this literal-attestation design and may require stronger phonetic or generative audio grounding. This bounds the available headroom for AVERT's current add and swap operators.

\paragraph{Training data.} Our base model uses the SpokenWOZ training split together with Loquacious \citep{loquacious2024} and Fisher \citep{cieri2004fisher} for ASR pre-training.  SpokenWOZ and Loquacious are open, but Fisher requires an LDC license to reproduce the ASR pre-training stage.

\paragraph{Slot selection and schema transfer.} The per-operator slot subsets and the add threshold are tuned on the SpokenWOZ development set and fixed before test evaluation. The procedure transfers to any schema with labeled development data, but we do not evaluate zero-shot transfer, and the gains may depend on the SpokenWOZ slot inventory. Selecting subsets without target-domain labels is left to future work.

\paragraph{Differential error exposure.} AVERT recovers a value only when it appears somewhere in the cumulative transcript. Values that ASR mistranscribes consistently, which is more likely for uncommon names and for underrepresented accents, therefore fall outside what attestation can reach. We do not measure how the 79.0\% unattested share is distributed across speakers.

\bibliography{custom}

\newpage
\appendix
\section{Operator Slot Subsets}
\label{app:slots}
Each operator is enabled on a slot subset selected on the development set: a slot is included when enabling it corrects more errors than it introduces. The subsets were fixed before any test evaluation and cover 27 of the 35 candidate slots. A slot may appear under more than one operator, but only one fires per turn, since each operator is conditioned on whether the slot is present, absent, or unattested.

\begin{table}[h]
\centering
\footnotesize
\setlength{\tabcolsep}{4pt}
\begin{tabular}{@{}l>{\raggedright\arraybackslash}p{0.72\columnwidth}@{}}
\toprule
Operator & Enabled slots \\
\midrule
\textsc{Vote} (12) & attraction-area, attraction-name, hotel-name, hotel-people, restaurant-pricerange, taxi-departure, taxi-leave, train-arrive, train-departure, train-destination, train-leave, train-people \\
\addlinespace[3pt]
\textsc{Add} (17) & attraction-type, hotel-area, hotel-internet, hotel-parking, hotel-people, hotel-stars, hotel-stay, hotel-type, profile-phonenumber, restaurant-area, restaurant-day, restaurant-name, restaurant-people, restaurant-pricerange, train-day, train-departure, train-destination \\
\addlinespace[3pt]
\textsc{Swap} (5) & hotel-stars, profile-name, restaurant-day, restaurant-food, train-destination \\
\bottomrule
\end{tabular}
\caption{Slots enabled for each operator, selected on the development set.}
\label{tab:slot-subsets}
\end{table}



\end{document}